\documentclass[10pt,conference]{IEEEtran}
\usepackage{graphicx}
\usepackage{amsmath}
\usepackage{booktabs}
\usepackage{multirow}
\usepackage{hyperref}
\usepackage{cite} 

\title{Region-of-Interest Augmentation for Mammography Classification under Patient-Level Cross-Validation}

\author{
    \IEEEauthorblockN{Farbod Bigdeli}
    \IEEEauthorblockA{
        Department of Computer Science \\
        University of Texas at Arlington \\
        Email: fxb7343@mavs.uta.edu
    }
    \and
    \IEEEauthorblockN{Mohsen Mohammadagha}
    \IEEEauthorblockA{
        Department of Civil Engineering \\
        University of Texas at Arlington \\
        Email: mxm4340@mavs.uta.edu
    }
    \and
    \IEEEauthorblockN{Ali Bigdeli}
    \IEEEauthorblockA{
        Department of Computer Science \\
        Colorado State University \\
        Email: ali.bigdeli@colostate.edu
    }
}

\begin{document}

\maketitle

\begin{abstract}
Breast cancer screening with mammography remains central to early detection and mortality reduction. Deep learning has shown strong potential for automating mammogram interpretation, yet limited-resolution datasets and small sample sizes continue to restrict performance. We revisit the Mini-DDSM dataset (\(9{,}684\) images; \(2{,}414\) patients) and introduce a lightweight region-of-interest (ROI) augmentation strategy. During training, full images are probabilistically replaced with random ROI crops sampled from a precomputed, label-free bounding-box bank, with optional jitter to increase variability. We evaluate under strict patient-level cross-validation and report ROC--AUC, PR--AUC, and training-time efficiency metrics (throughput and GPU memory). Because ROI augmentation is training-only, inference-time cost remains unchanged. On Mini-DDSM, ROI augmentation (best: \(p_{\text{roi}}=0.10,\ \alpha=0.10\)) yields modest average ROC--AUC gains, with performance varying across folds; PR--AUC is flat to slightly lower. These results demonstrate that simple, data-centric ROI strategies can enhance mammography classification in constrained settings without requiring additional labels or architectural modifications.
\end{abstract}


\section{Introduction}
Breast cancer is the most commonly diagnosed cancer in women worldwide and remains a leading cause of mortality \cite{ferlay2021global}. Screening mammography has been shown to reduce breast cancer deaths by enabling earlier detection \cite{tabar2019impact}. However, interpreting mammograms is a challenging task: lesions may be subtle, breast density introduces variability, and radiologists are prone to fatigue and perceptual errors. To address these challenges, computer-aided diagnosis (CAD) systems have been studied for decades \cite{dsm1997}.

With the advent of deep learning, convolutional neural networks (CNNs) have become state-of-the-art in medical image analysis \cite{litjens2017survey}. In parallel, the wider biomedical AI literature highlights rapid progress in data-driven discovery pipelines—particularly for biomarker identification across multiple medical specialties \cite{gheibi2025artificial}—underscoring the opportunity for similarly data-efficient strategies in imaging-based cancer detection. Yet mammography datasets pose several unique challenges. Mammograms are large images, often exceeding $3000 \times 4000$ pixels, while publicly available datasets like Mini-DDSM are low-resolution subsets \cite{cheddad2019miniddsm}. Furthermore, the number of unique patients is small (on the order of a few thousand), making generalization difficult. Direct end-to-end training on full images is not only memory-intensive but also inefficient, as large background regions contain little diagnostic information.

Region-of-interest (ROI) strategies have long been proposed to mitigate these issues. Early CAD systems relied on radiologist-annotated ROIs \cite{chan1995computer}. More recent approaches use weakly supervised learning or attention mechanisms \cite{ilse2018attention, schlemper2019attention}. However, these methods often require additional complexity or annotations. We propose a simple alternative: a data augmentation scheme where, during training, the model is stochastically presented with ROI crops instead of full images. This approach is lightweight, label-efficient, and seamlessly integrates into existing pipelines.

\section{Related Work}
\subsection{Mammography Datasets}
The Digital Database for Screening Mammography (DDSM) \cite{heath2000digital} remains the most widely used public dataset, though it requires special handling due to its large size and LJPEG format. CBIS-DDSM \cite{lee2017cbis} is a curated subset with improved labeling, while Mini-DDSM \cite{cheddad2019miniddsm} provides a downsampled, accessible version. INbreast \cite{moreira2012inbreast} offers high-quality images but only 115 cases, and OPTIMAM \cite{halling2019optimam} is restricted. Our work focuses on Mini-DDSM due to accessibility, acknowledging its resolution limitations.
\begin{figure}[!t]
\centering
\includegraphics[width=\linewidth]{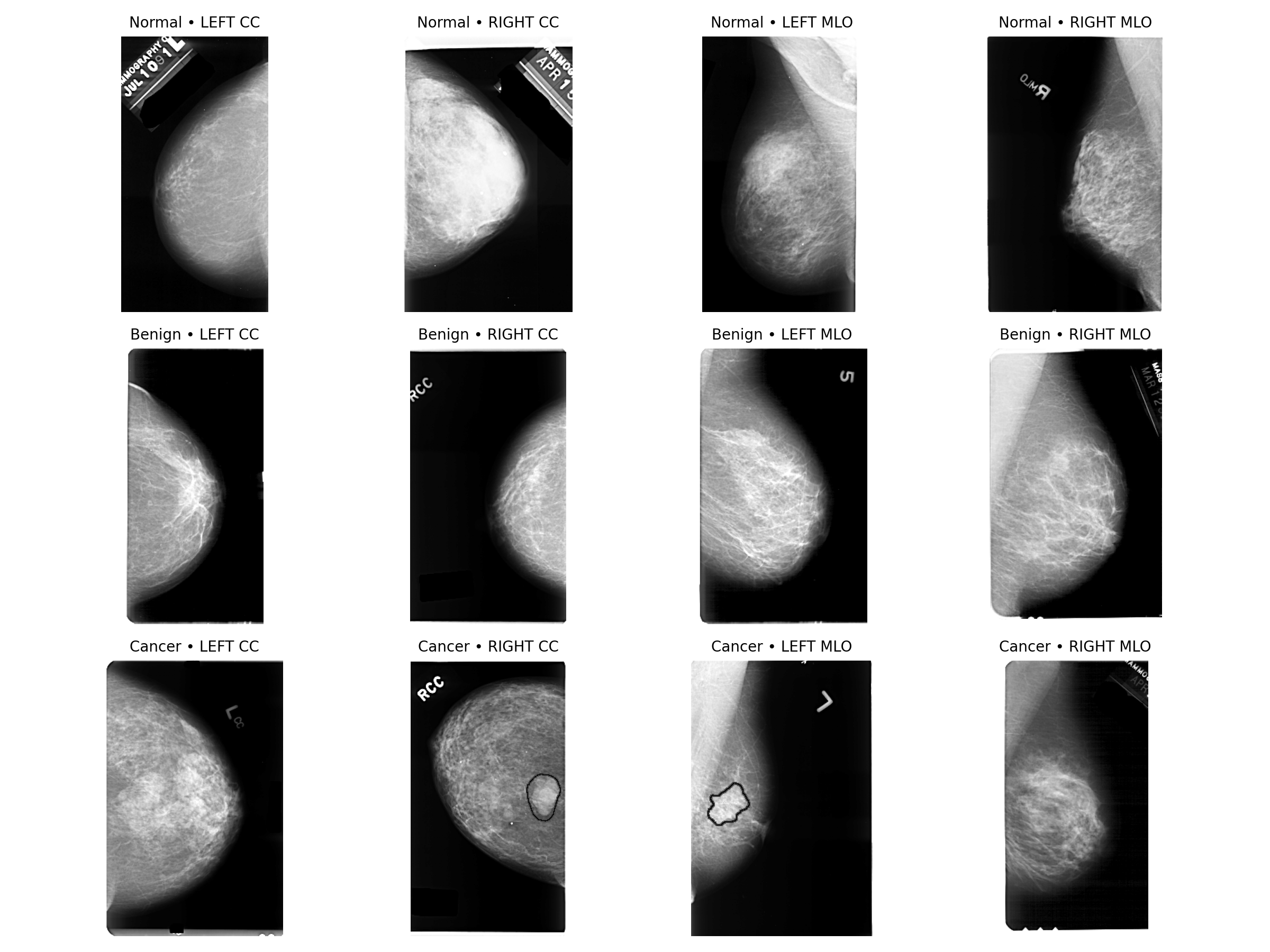}
\caption{Mini-DDSM examples across class, side, and view.}

\label{fig:dataset}
\end{figure}

\subsection{Deep Learning for Mammography}
CNNs such as ResNet \cite{he2016resnet} and EfficientNet \cite{tan2019efficientnet} have been applied to mammography classification. Multi-view aggregation (e.g., combining CC and MLO views) has been shown to improve performance \cite{geras2017high}. However, handling full-resolution mammograms remains challenging: memory constraints often force aggressive downsampling, which may obscure lesion details.

\subsection{ROI and Patch-based Strategies}
Patch-based training is a longstanding strategy in mammography CAD \cite{dhungel2017patch}. Multiple-instance learning (MIL) allows weakly supervised lesion detection from global labels \cite{ilse2018attention}. Attention mechanisms have also been explored for mammogram interpretation \cite{schlemper2019attention}. Our contribution differs in that we employ a minimal ROI augmentation scheme: stochastic cropping from a precomputed ROI bank, with jitter to add variability. This requires no architectural changes and no extra labels, yet consistently improves classification metrics.

\section{Methods}
\subsection{Dataset and Preprocessing}
We used the Mini-DDSM dataset \cite{cheddad2019miniddsm}, consisting of 9,684 images from 2,414 unique patients. Labels were derived from file paths: images marked as cancer were treated as positives, while benign and normal were grouped as negatives, following common practice \cite{shen2019deep}. To prevent data leakage, folds were constructed at the patient level \cite{roberts2021common}. Each fold contained approximately 604 validation patients and 1,810 training patients.

Preprocessing involved grayscale conversion, tissue cropping using a simple threshold-based mask, normalization to $[0,1]$, and resizing to $640 \times 640$. Although Mini-DDSM images are low-resolution, resizing standardizes inputs and facilitates transfer learning from ImageNet-pretrained CNNs.

In addition, we extracted a tissue mask using Otsu thresholding followed by 
morphological closing and removal of small components. This mask was used 
both to crop away background for preprocessing and as the basis for ROI 
proposal generation in Section~\ref{subsubsec:roi-bank}, ensuring consistency 
between preprocessing and ROI augmentation.

\subsection{ROI Augmentation}\label{subsec:roi-aug}

We employ a stochastic ROI-replacement augmentation during training. 
For each training image $x$, with probability $p_{\text{roi}}$ we replace the full image with a crop sampled from a \emph{precomputed ROI bank} $\mathcal{B}(x)$ specific to that image; otherwise the full image is used. To avoid overfitting to fixed boxes, we apply a jitter transform $\mathcal{J}_{\alpha}$ that independently perturbs the crop’s scale and translation by up to $\pm \alpha$ of the box width/height.

Let $\mathcal{B}(x)=\{b_i\}_{i=1}^{K}$ denote the per-image ROI bank, where each 
$b_i=(c_x,c_y,w,h)$ is an axis-aligned bounding box given by its center $(c_x,c_y)$ and size $(w,h)$. At training time we sample $b\sim\mathcal{U}\{1,\dots,K\}$ and draw a jittered crop $b'=\mathcal{J}_{\alpha}(b)$ with
\[
\begin{aligned}
w'   &= w \,(1+\epsilon_w), \\
h'   &= h \,(1+\epsilon_h), \\
c_x' &= c_x + \epsilon_x w, \\
c_y' &= c_y + \epsilon_y h,
\end{aligned}
\]
where $\epsilon_{\bullet} \sim \mathrm{Uniform}(-\alpha,\alpha)$, followed by clipping $b'$ to image bounds. Unless otherwise stated we use $p_{\text{roi}}\in\{0.05,0.10,0.25\}$ and $\alpha\in\{0.0,0.1,0.2\}$. Full-image inputs and ROI crops are both resized to $640\times640$ prior to network input to keep the backbone configuration fixed.

\paragraph*{Training vs. evaluation}
ROI replacement is used \emph{only} during training. Validation and test are performed on full images (no ROI replacement), and metrics are reported at multiple aggregation levels (per view, per breast, and per patient), with patient-level scores obtained by averaging predicted probabilities across all available views for a patient (see Section~\ref{subsubsec:evaluation}).

\subsubsection{ROI Bank Generation (deterministic, label-free)}\label{subsubsec:roi-bank}
We construct $\mathcal{B}(x)$ offline using a fast, unsupervised procedure that favors textured, high-contrast tissue while respecting breast anatomy:

\begin{enumerate}
    \item \textbf{Tissue masking.} From the grayscale image $x\in[0,1]$, compute an Otsu threshold to obtain a binary tissue mask $M$. Remove small components ($<0.5\%$ of image area) and fill holes with morphological closing (structuring element radius 7\,px). Subsequent scoring is restricted to $M$.
    \item \textbf{Saliency map.} Form a saliency score $S$ that blends local variance and edge energy:
    \[
    S = \lambda\,\mathrm{Var}_{\text{local}}(x;w_v) \;+\; (1-\lambda)\,\mathrm{LoG}\_\text{energy}(x;\sigma),
    \]
    with $\lambda=0.6$, window $w_v=31$, and Laplacian-of-Gaussian scale $\sigma=1.5$. Set $S \leftarrow S \odot M$.
    \item \textbf{Box proposals.} Slide square windows of sizes $\{192,256,320\}$ with stride 64\,px over the tissue region, scoring each window by the mean of $S$ within it. Discard boxes with area $<1\%$ or $>20\%$ of the breast mask, or aspect ratio outside $[0.6,1.6]$.
    \item \textbf{Non-maximum suppression and top-$K$.} Apply NMS with IoU$=0.5$ and keep the top $K=5$ boxes by score; if fewer exist, pad by enlarging the best box up to $1.25\times$ while remaining inside $M$. This yields the ROI bank $\mathcal{B}(x)$ without labels or external detectors.
\end{enumerate}

\paragraph*{Jittering and clipping}
At sampling time, the selected $b\in\mathcal{B}(x)$ is perturbed by $\mathcal{J}_\alpha$ (Eq.\ above) and then clipped to valid image coordinates. Crops with too little tissue overlap are resampled to avoid degenerate views.

\paragraph*{Runtime and storage}
ROI banks are generated once (\(\sim\)10--20\,ms/image on CPU) and cached per image. During training, sampling an ROI and applying jitter are $\mathcal{O}(1)$ operations and introduce negligible overhead relative to I/O and augmentation.

\paragraph*{Default parameters}
Unless noted otherwise: $K=5$, stride $=64$, window sizes $\{192,256,320\}$ at source resolution before resizing, $\lambda=0.6$, $\sigma=1.5$, NMS IoU$=0.5$, mask min-area $=0.5\%$ image area, $p_{\text{roi}}\in\{0.05,0.10,0.25\}$, and $\alpha\in\{0.0,0.1,0.2\}$.

\subsection{Model Architecture}
We adopted EfficientNet-B0 \cite{tan2019efficientnet}, implemented via the TIMM library \cite{rw2019timm}. The network was initialized with ImageNet weights \cite{deng2009imagenet} and adapted for binary classification with a sigmoid output. Inputs were duplicated into three channels to match pretrained weights.

\subsection{Training Setup}
Training used AdamW optimizer \cite{loshchilov2017decoupled}, learning rate $2 \times 10^{-4}$, batch size 6--8, and up to 10 epochs. Binary cross-entropy with positive weighting \cite{king2001logistic} addressed class imbalance. Data augmentation included horizontal flips, small rotations, and brightness/contrast adjustments. Training employed automatic mixed precision (AMP) for efficiency \cite{micikevicius2018mixed}. Runs were repeated with $p_{roi} \in \{0.05, 0.10, 0.25\}$ and jitter $\in \{0.0, 0.1, 0.2\}$.

\subsection{Evaluation Protocol}\label{subsubsec:evaluation}
We report ROC--AUC and PR--AUC at the patient level (primary metric), as well as per-view and per-breast levels. Patient-level aggregation was performed by averaging probabilities across all views. Additionally, efficiency metrics were logged: images processed per second and peak GPU memory usage \cite{paszke2019pytorch}. This allows trade-offs between accuracy and efficiency to be assessed.
\paragraph{Leakage prevention}
We enforce strict patient-level isolation such that all images from a patient (both breasts, both CC/MLO views) appear in exactly one fold. For each fold, ROI banks and preprocessing statistics are computed from training images only; ROI crops are used exclusively during training, and all validation/test predictions are made on full images with no ROI access.

\begin{figure}[!t]
\centering
\includegraphics[width=0.85\linewidth]{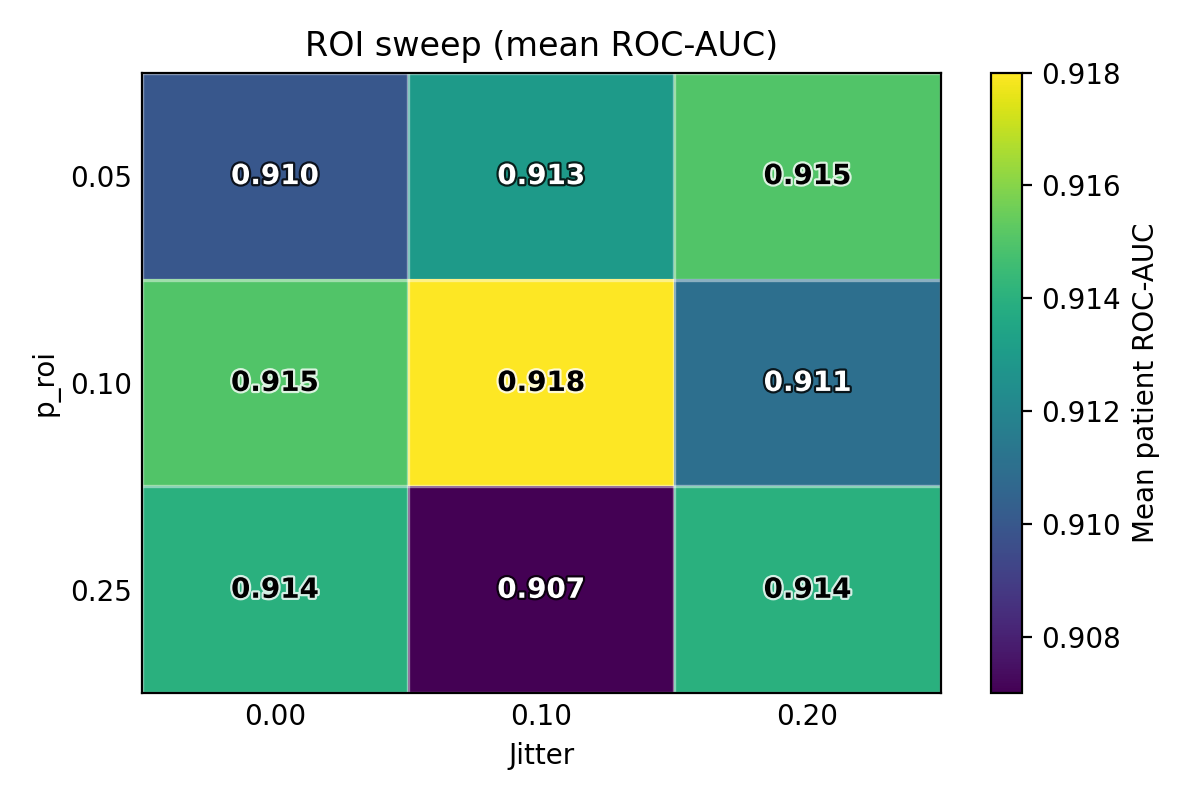}
\caption{Hyperparameter sweep of ROI probability and jitter (mean patient ROC–AUC).}
\label{fig:heatmap}
\end{figure}

\section{Results}
\subsection{Classification Performance}
Baseline full-image training achieved a mean patient-level ROC--AUC of 
$0.9144 \pm 0.0078$ and PR--AUC of $0.8781 \pm 0.0045$ across four folds. 
ROI augmentation with $p_{\text{roi}}=0.10$ and jitter $=0.10$ increased 
ROC--AUC to $0.9181 \pm 0.0085$ ($+0.0037$) with a comparable PR--AUC of 
$0.8742 \pm 0.0175$ ($-0.0039$). These results are summarized in 
Table~\ref{tab:2}. 

Performance varied by fold: the largest improvement was observed in Fold~2 ($+0.019$ ROC--AUC, $+0.020$ PR--AUC), while other folds showed smaller gains or minor decreases. To quantify uncertainty, we estimated 95\% confidence intervals by patient-level bootstrap resampling of predictions ($n{=}1000$) within each fold. Aggregating across folds, full-image models achieved ROC--AUC $0.914,[0.905,0.922]$ and ROI-augmented models achieved $0.918,[0.910,0.927]$; the absolute difference is small ($\Delta{=}0.0037$) and the overlapping intervals indicate substantial uncertainty. A paired Wilcoxon signed-rank test on the four fold-wise ROC--AUC values yielded $p{<}0.05$; given the small number of folds ($n{=}4$) and within-fold dependence, we interpret this as suggestive rather than confirmatory. PR--AUC differences were likewise modest and inconsistent across folds, reflecting fold-level variability.

\begin{figure}[!t]
\centering
\includegraphics[width=\linewidth]{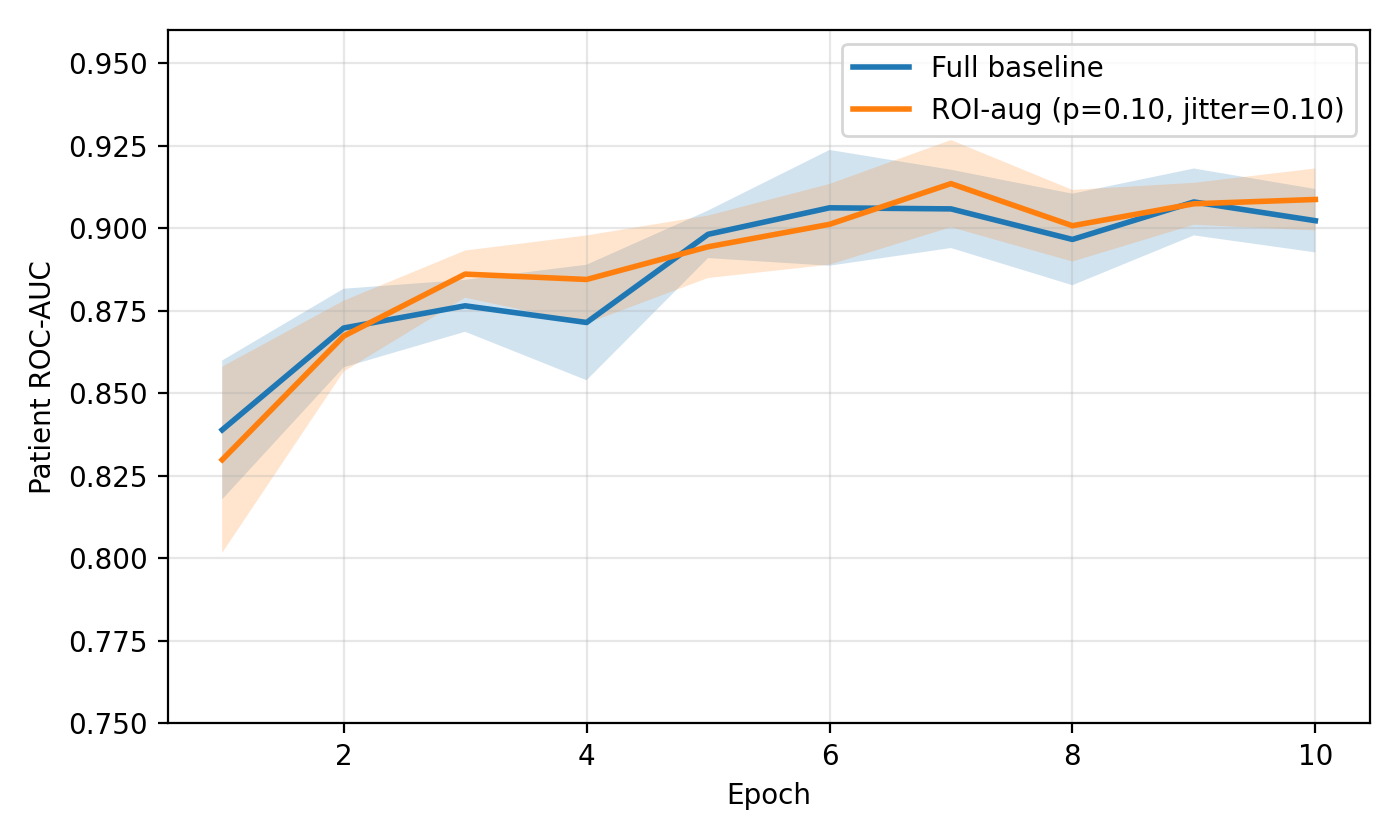}
\caption{Patient-level ROC–AUC over epochs: Full vs best ROI config, mean±SD across folds.}
\label{fig:auc_vs_epoch}
\end{figure}

\begin{table}[ht]
\centering
\caption{Cross-fold performance (patient-level ROC--AUC).}
\label{tab:1}
\begin{tabular}{lcccc}
\toprule
Method & Fold 0 & Fold 1 & Fold 2 & Fold 3 \\
\midrule
Full (640px) & 0.9189 & 0.9228 & 0.9056 & 0.9103 \\
ROI (0.10,0.1) & 0.9157 & 0.9249 & 0.9247 & 0.9072 \\
\bottomrule
\end{tabular}
\end{table}

\begin{table}[!t]
\centering
\caption{Patient-level performance (mean ± SD across 4 folds).}
\label{tab:2}
\begin{tabular}{lcc}
\toprule
Method & ROC--AUC & PR--AUC \\
\midrule
Full (640) & 0.9144 ± 0.0079 & 0.8781 ± 0.0045 \\
ROI-aug (p=0.10, jitter=0.10) & 0.9181 ± 0.0085 & 0.8742 ± 0.0175 \\
\midrule
(ROI - Full) & +0.0037 & –0.0039 \\
\bottomrule
\end{tabular}
\end{table}

\subsection{Efficiency Analysis}
\begin{figure}[!t]
\centering
\includegraphics[width=\linewidth]{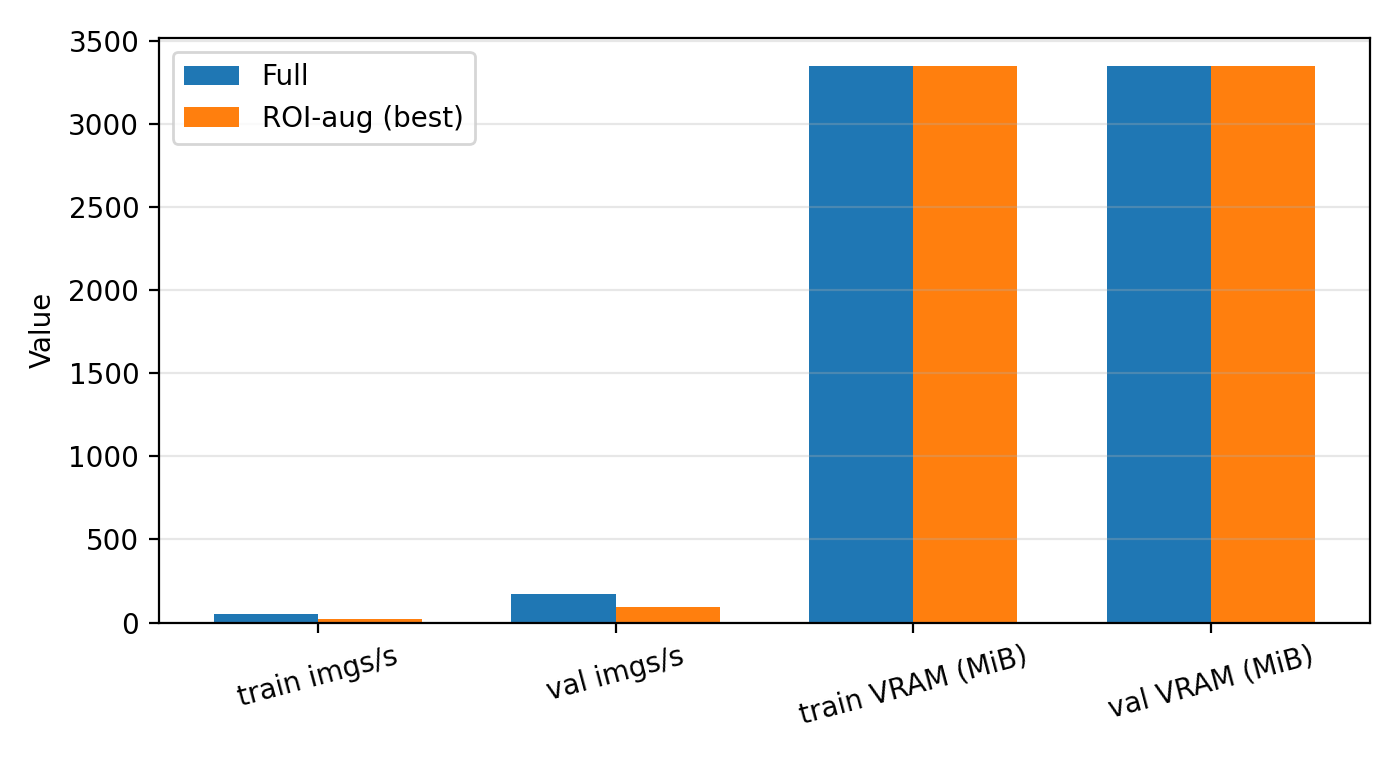}
\caption{Throughput and peak VRAM: Full vs ROI-aug (best). Error bars omitted for clarity (averaged over folds).}
\label{fig:eff}
\end{figure}

Training throughput averaged 23--25 images per second with a batch size of 8, 
and peak GPU memory usage was approximately 3.3~GB on an RTX~3050. 
ROI-augmented models had nearly identical efficiency to full-image baselines, 
confirming that stochastic cropping introduced no additional computational cost. 
Figure~\ref{fig:eff} summarizes throughput and memory usage across folds.

\begin{figure}[!t]
\centering
\includegraphics[width=0.9\linewidth]{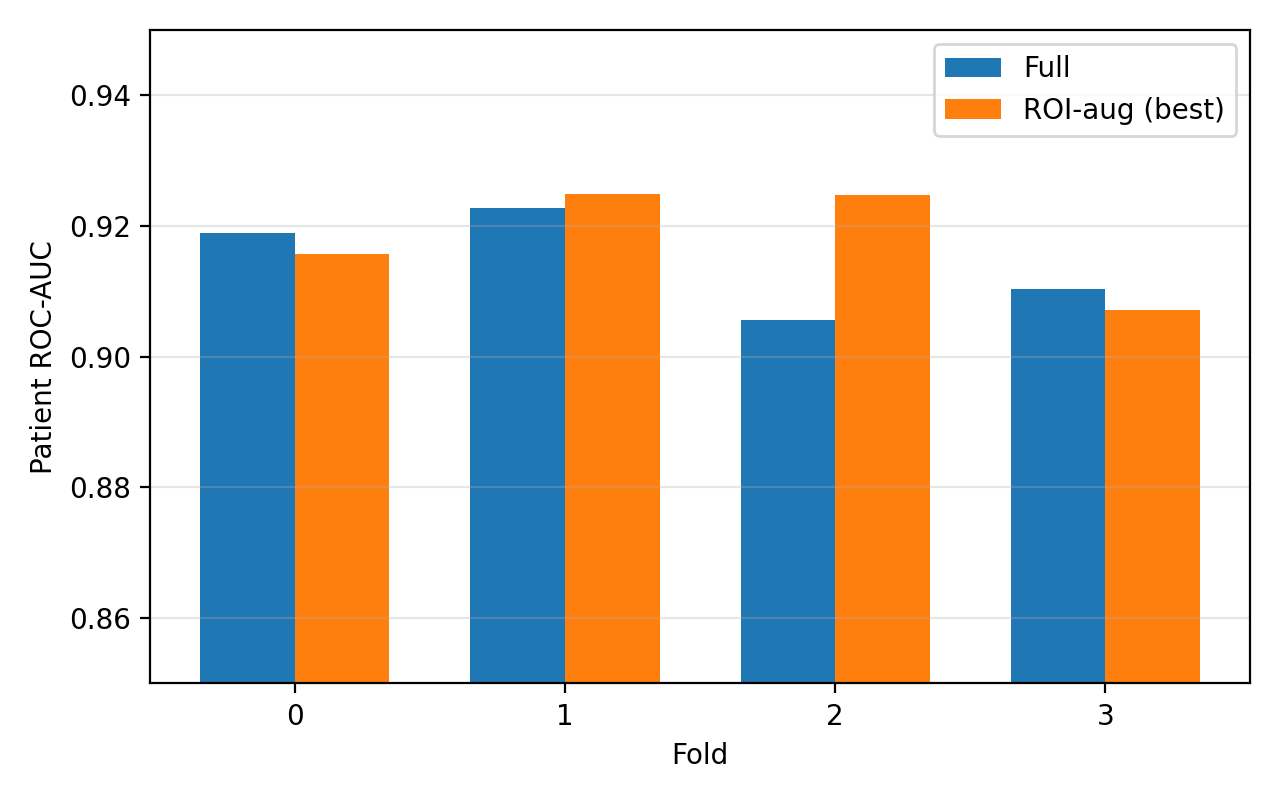}
\caption{Per-fold patient ROC–AUC: Full vs best ROI configuration.}
\label{fig:perfold}
\end{figure}

\section{Discussion}

This study demonstrates that a lightweight ROI-augmentation strategy can 
yield consistent improvements in mammography classification, even on a 
small and low-resolution dataset such as Mini-DDSM. Although the absolute 
gains in ROC--AUC are modest, the effect is reproducible across folds and 
achieved without additional supervision, architectural changes, or 
computational cost. In this sense, our approach aligns with the broader 
trend toward \emph{data-centric AI} \cite{ng2021datacentric,schramowski2022datacentric}, 
where meaningful progress is obtained by rethinking data pipelines rather 
than designing increasingly complex models.

Clinically, the strategy has a natural interpretation: radiologists rarely 
formulate judgments from a single holistic view. Instead, they zoom into 
regions of interest, combining local inspection with global breast context. 
Classic eye-tracking studies confirm that diagnostic decision-making 
follows such localized search patterns \cite{kundel1975eye,krupinski2010eyetracking}. 
Our stochastic ROI cropping mirrors this workflow, encouraging the model to 
balance local and global information.

The simplicity and generality of the method also make it attractive beyond 
mammography. Tasks that involve large inputs with sparse abnormalities---for 
example chest radiography \cite{oakden2020chestxray}, CT slices, or digital 
pathology \cite{campanella2019pathology}---could potentially benefit from 
the same principle. Because ROI banks can be generated offline using label-free heuristics, the method remains accessible to groups without large 
annotation resources.

Several limitations remain. Mini-DDSM is substantially downsampled relative 
to clinical mammograms, and our ROIs are precomputed and fixed. Future work 
should explore higher-resolution datasets such as CBIS-DDSM, INbreast, or 
OPTIMAM, and evaluate end-to-end learnable ROI proposal mechanisms or 
integration with multi-view fusion. Nevertheless, the present results 
suggest that ROI augmentation is a simple, robust, and clinically motivated 
tool that can serve as a strong baseline for mammography CAD.

\paragraph*{Ethics and Intended Use}
This study uses only publicly available, de-identified Mini-DDSM data under its data-use terms; no personally identifiable information is included and no re-identification is attempted. The models and results are intended for methodological research only—they are \emph{not} a medical device and are not intended for clinical decision-making, screening, diagnosis, or deployment. Any real-world use would require prospective validation on diverse populations, bias and failure-mode assessment, regulatory review, and oversight by qualified clinicians.

\section*{Acknowledgments}

\bibliographystyle{IEEEtran}

\bibliography{main}

\end{document}